\documentclass{article}

\usepackage[preprint]{neurips_2026}
\workshoptitle{Machine Learning for Systems Workshop at NeurIPS 2026}

\usepackage{wrapfig}

\usepackage[utf8]{inputenc}
\usepackage[T1]{fontenc}
\usepackage{hyperref}
\usepackage{url}
\usepackage{booktabs}      
\usepackage{amsfonts}
\usepackage{nicefrac}
\usepackage{microtype}
\usepackage{xcolor}
\usepackage{graphicx}     

\title{Argus: A Real-EKS Study of When Predicting Spot Interruptions Beats Simple Checkpointing}

\author{
  Angshuman Chakravertty\\
  SVKM's NMIMS (Deemed to be University)\\
  \texttt{angshumanchakravertty2@gmail.com}
  \And
  MD Rayyan\\
  SVKM's NMIMS (Deemed to be University)\\
  \texttt{rayyan1652@gmail.com}
}

\begin{document}
\maketitle

\begin{abstract}

Elastic Compute Cloud (EC2) Spot is 60\% to 90\% cheaper than On-Demand but can be reclaimed on just a 2-minute notice; for expensive multi-node training this loss can be severe, with one reclaim costing hours of synchronous progress. We build Argus, a Kubernetes operator, and ask empirically, on a CIFAR-10 testbed, when predicting interruptions beats simple checkpointing. Argus on real EKS survives a real Spot drain with a graceful SIGTERM checkpoint, resuming from epoch 8 and losing only the in-progress epoch. Alongside, we further find that in an 80-trial benchmark, the reactive-on-notice degrades toward no protection once interruption outpaces the fixed 2-minute notice, and predictive wasted compute is driven to zero, but with an oversized fixed lead it over-migrates so severely that at the fastest rate only one of five runs completes, while periodic is a strong ML-free baseline. A lead-time sweep turns the lead prediction into a guideline where a small lead suffices for zero waste, but excess lead is wasteful. The predictor built is advisory (a proxy label); real interruption labels and large-model-scale validation are future work.

\end{abstract}

\section{Introduction}
\label{sec:intro}

AWS Elastic Compute Cloud (EC2) Spot is 60\% to 90\% cheaper than On-Demand but only has a 2-minute warning window and can be reclaimed~\citep{aws-spot}. A simple reclaim's loss is trivial for a cheap job, but when it comes to expensive multi-node training~\citep{megatron-lm}, the loss is severe, where several hours of synchronous multi-GPU progress get discarded. At illustrative On-Demand list prices~\citep{aws-spot}, a 16-node A100 job (128 GPUs) at roughly \$12/node-hr Spot, checkpointing hourly, loses on the order of \$190 of synchronous compute to a single node's reclaim, against a checkpoint cost of cents, three orders of magnitude. During this work CIFAR-10 was used as a cheap, controlled testbed, with the mechanism being workload-agnostic; no large-model experiments are claimed, and multi-node validation is future work.

The usefulness of the notice depends on the interruption frequency, and exposure compounds with scale. For an $N$-node job in which each node has per-interval reclaim probability $p$, $P(\geq 1) = 1 - (1-p)^N$ is the chance that at least one node is reclaimed and rises steeply with $N$ (at $p=0.05$, $N=16$, about 56\% per interval). GPU Spot pools have the highest interruption-frequency tiers~\citep{aws-spot-advisor}, thus a large multi-node job faces interruptions far more often than any single node, pushing it into a regime where there is no useful lead found from a fixed 2-minute reactive notice. Prior systems make training on transient resources preemptible~\citep{varuna, bamboo}; therefore, instead, we ask \emph{when} the free reactive notice is beaten by a predictive checkpoint policy and validate it on real Elastic Kubernetes Service (EKS). We build Argus, a Kubernetes operator, to answer the question empirically on the controlled testbed.

In this work we make three contributions: (1) a real-EKS-validated interruption-survival result (Section~\ref{sec:realeks}), (2) a controlled benchmark isolating when predictive checkpointing beats periodic and reactive-on-notice (Section~\ref{sec:predpay}), (3) a lead-time sensitivity result showcasing how much lead prediction needs to turn into a concrete guideline (Section~\ref{sec:leadtime}).

\section{System Design}
\label{sec:design}

\begin{figure*}[h]
  \centering
  \includegraphics[width=\linewidth]{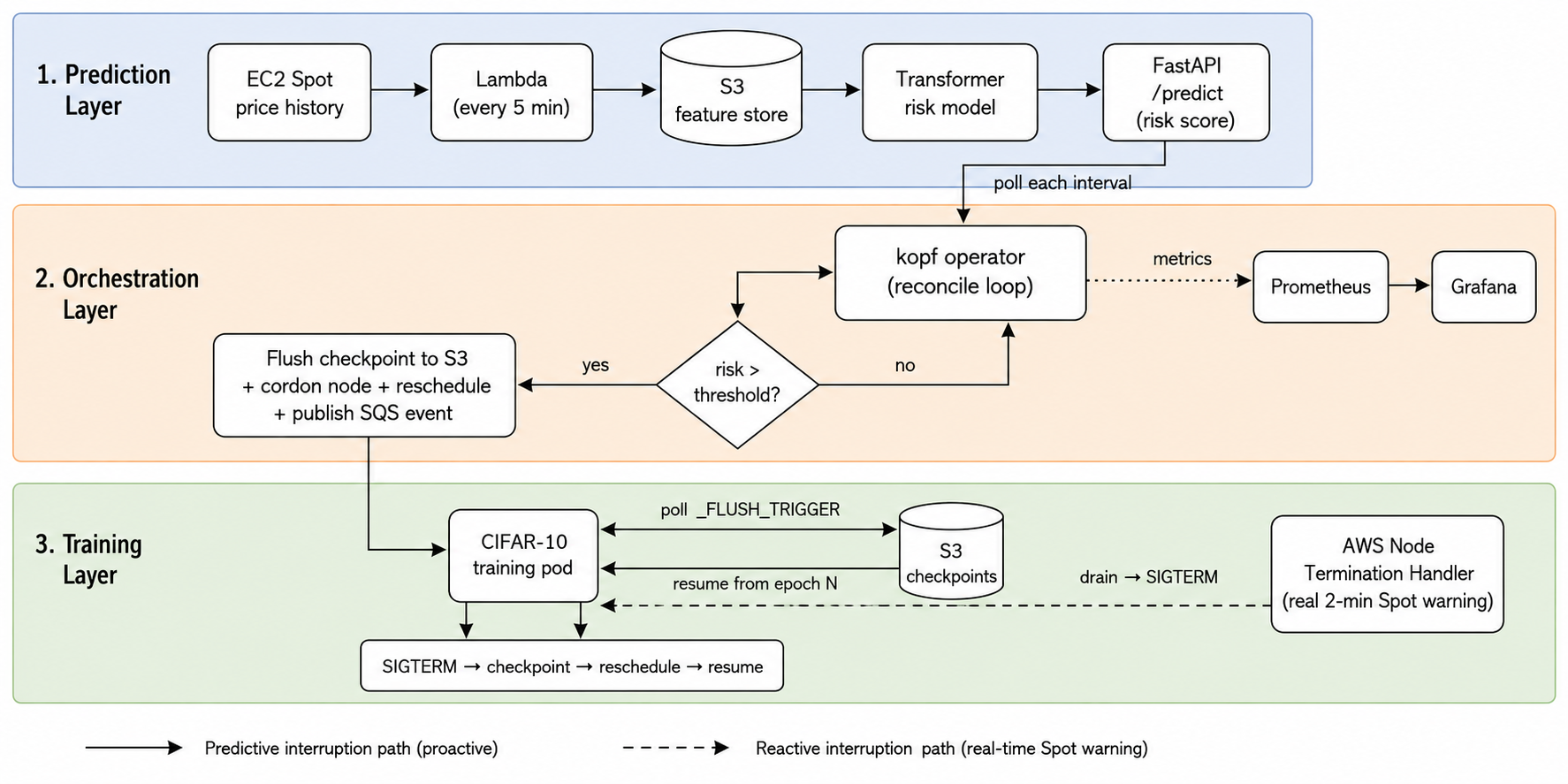}
    \caption{Argus architecture. Two triggers (predictive, when the risk score crosses the threshold; reactive, the real 2-minute Spot warning via NTH) drive one SIGTERM, checkpoint, reschedule, resume flow.}
  \label{fig:arch}
\end{figure*}

Argus is composed of three primary layers (Figure~\ref{fig:arch}): a prediction layer, in which Lambda pulls Spot price history every 5 minutes into an S3 feature store and then a Transformer risk model is served via FastAPI (\texttt{/predict}); an orchestration layer, consisting of a \texttt{kopf} Kubernetes operator~\citep{k8s-operators} with a \texttt{SpotResilientJob} (Custom Resource Definition) CRD alongside a reconcile loop, which polls \texttt{/predict} each interval; and a training layer, where the job resumes after persisting checkpoints to S3. During a high-risk decision, the operator only writes a \texttt{\_FLUSH\_TRIGGER} marker to S3 and cordons the node and reschedules the job; thus, the checkpoint mechanism is intentionally left decoupled. The training pod owns the actual flush of \texttt{model.pt} by polling the marker because the operator never touches the training internals and keeps it workload-agnostic.

A pivotal decision involved having two triggers drive the same SIGTERM, checkpoint, reschedule, and resume path: a predictive trigger, where the operator pre-migrates ahead of any notice when the risk score crosses the threshold, and a reactive trigger, where AWS's real 2-minute Spot warning via the node termination handler (NTH)~\citep{aws-nth} drains the node, which then delivers a SIGTERM, and the pod's handler checkpoints. EKS alongside IAM Roles for Service Accounts (IRSA) utilizes OpenID Connect (OIDC) to give pods AWS credentials with zero static keys.

\section{Evaluation}

\subsection{Surviving a real Spot drain on real EKS}
\label{sec:realeks}

The deployment was run on real EKS, real Spot nodes (\texttt{c5.xlarge/m5.xlarge})~\citep{aws-spot}, IRSA with zero static credentials, and AWS Node Termination Handler~\citep{aws-nth} in queue mode. When the training was live and at epoch 7, the interruption was fired, and NTH received the \texttt{EC2 Spot Instances Interruption Warning}, drain then evicted the pod with graceful SIGTERM, and then the SIGTERM handler wrote the checkpoint (epoch 8) to S3, as seen in Figure~\ref{fig:s3ckpt}. The node was cordoned and drained in 10 seconds, and the replacement pod on the healthy node showed "Resuming from epoch 8" and only the in-progress epoch's work was lost. The entire timeline of the process can be observed in Table \ref{tab:obj1}.

The interruption that was injected was schema-conformant (\texttt{EC2 Spot Instance Interruption Warning}) and not an AWS-issued Fault Injection Service (FIS) reclaim~\citep{aws-fis}. Nevertheless, NTH cannot distinguish injected from real, and thus the flow of drain, SIGTERM, checkpoint, and resume path is genuinely exercised, and we hope to include FIS-forced reclaim as part of the future development, because it was prevented on an account-subscription problem rather than the design.

\subsection{When does prediction pay?} 
\label{sec:predpay}

We compare four checkpoint policies on a shared synthetic training job:
\emph{no-protection}, \emph{periodic} (fixed-interval, signal-unaware),
\emph{reactive-on-notice} (a 2-minute notice, the honest baseline to
beat), and \emph{predictive}, which checkpoints on the model's risk
signal ahead of the notice. A harness kills the job on a Poisson process
at four mean time between failures (MTBF) rates (120, 300, 600, 1800\,s), five repetitions each, 80 trials total, injected locally, which isolates policy behavior from Spot-market variability and is cheap enough for the full sweep. An early harness race let a respawned process be killed before its checkpoint confirmed (24:1 respawn-to-checkpoint before the fix, 1:1 after); we verified before trusting Table~\ref{tab:bench}.

\begin{table}[t]
  \centering
  \caption{Results at the fastest tested rate (MTBF = 120\,s, 5 repetitions).}
  \label{tab:bench}
  \small
  \begin{tabular}{lrrrr}
    \toprule
    Arm & Compl. & Wasted (s) & Makespan (s) & Ckpts \\
    \midrule
    No-protection               & 1.0 & $202.3 \pm 141.1$ & 361.8 & 0.0 \\
    Reactive-on-notice          & 1.0 & $169.4 \pm 105.8$ & 328.0 & 4.8 \\
    Periodic                    & 1.0 & $4.3 \pm 3.4$     & 158.8 & 24.0 \\
    \textbf{Predictive (Argus)} & \textbf{0.2} & $\mathbf{0.0 \pm 0.0}$ & \textbf{353.2} & \textbf{221.6} \\
    \bottomrule
  \end{tabular}
\end{table}

Table~\ref{tab:bench}'s fastest-rate row makes the case directly: wasted
compute is 202.3\,s for no-protection, 169.4\,s for reactive, 4.3\,s for
periodic, and 0.0\,s for predictive, and Figure~\ref{fig:waste} shows this
pattern pooled across all four rates. Reactive's waste collapses toward
no-protection's because its 120\,s notice roughly equals this rate's mean
interval, so each notice fires with the next interruption already close
behind, leaving too little lead to checkpoint; more generally, a free
2-minute notice stops helping once interruptions arrive faster than
about once every two minutes, a regime not exotic for the large,
multi-node fleets in Section~\ref{sec:intro}.
Predictive's 0.0\,s here assumes a benchmark-parameter lead time, not a
measured model property (Section~\ref{sec:leadtime}), while periodic,
with no learned signal at all, nearly matches it on wasted compute and
wins decisively on both makespan (158.8\,s vs.~ 353.2\,s) and checkpoint
count (24 vs.~ 221.6), because at this rate predictive's 600\,s lead
badly exceeds the interval and only one of five runs completes cleanly
(Table~\ref{tab:bench}), a failure Section~\ref{sec:leadtime} diagnoses.
Predictive's only genuinely clean win here is zero wasted compute; we
credit periodic's strength openly rather than round that up into a
larger claim.

\subsection{How much lead time does prediction need? (sensitivity)}
\label{sec:leadtime}

\begin{figure}[t]
  \centering
  \includegraphics[width=\linewidth]{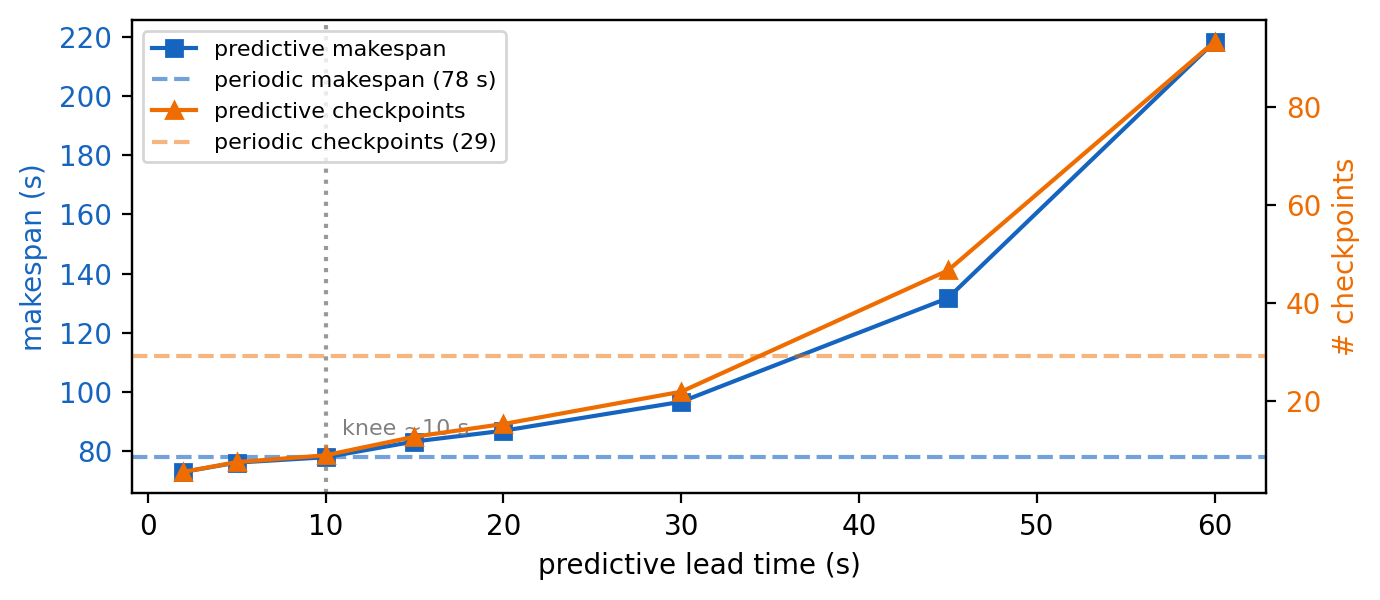}
  \caption{Lead-time sensitivity. Wasted compute stays at zero across all leads, while makespan and checkpoint count climb once the lead exceeds the interruption interval.}
  \label{fig:leadsweep}
\end{figure}

We swept the predictive arm's lead time across $\{2, 5, 10, 15, 20, 30,
45, 60\}$\,s at a fixed 20\,s mean interruption interval, five repetitions each,
against the ML-free periodic arm as reference (Figure~\ref{fig:leadsweep}).
Wasted compute is 0.0\,s at every lead, \textbf{even 2\,s}, since the
operator checkpoints synchronously before migrating, so a few seconds
suffices and the practical lower bound is the checkpoint write time.
Excess lead is not free, though: predictive's makespan stays at or below
periodic's (roughly 78\,s) only up to a knee near the interval, crossing
around 10--15\,s (78.0\,s at 10\,s, 83.3\,s at 15\,s) (full data in Table~\ref{tab:leadsweep}) before rising
steeply; at 60\,s, three times the interval, it migrates about 93 times
and makespan nearly triples (218\,s vs.~ 73\,s at 2\,s). Below the knee
it stays strictly better than periodic: zero waste against 4.3\,s,
lower-or-equal makespan, and far fewer checkpoints (5-9 vs.~ 29).

Lead time should be set just above the checkpoint write time and well
below the interval; beyond that, extra lead buys no further waste
reduction while linearly inflating overhead and makespan, exactly what
happened to predictive's 600\,s lead in Section~\ref{sec:predpay}. The synthetic checkpoint here is near-instant (in production, the real write time), and the model's 600 s lead errs long, so the fix is capping the operator's effective lead, not a bigger model.

\subsection{The risk model (brief)}
\label{sec:riskmodel}

A Transformer risk model~\citep{Transformer}, trained with focal loss~\citep{focal-loss} on a proxy label (Spot price spikes greater than 1\%, not real reclaims), required fixing a chain of concrete bugs to become real: a train/serve scaler skew (the old flat 0.0419 output), train/validation leakage, a focal-loss alpha no-op, and uncalibrated outputs. It shows a 14.68$\times$ base-rate lift (5-seed mean, 95\% CI $\approx$ [9.9$\times$, 19.5$\times$]) on that proxy label, while calibrated scores are tiny (best-F1 threshold $\approx$ 0.0015). The model is advisory only; its ceiling is label quality, and two features that should have helped did not (a cross-AZ feature and a real interruption-rate feature). We never claim it predicts real interruptions.

\subsection{Observability}

The operator emits three Prometheus metrics; predicted risk crossing the threshold triggers the checkpoint, and the proactive checkpoint firing at the same instant is shown by the provisioned Grafana dashboard (Figure~\ref{fig:grafana}). This was captured on the real operator against the real S3/SQS. 

\section{Limitations}

Argus consists of four main limitations:
(a) the interruption in Section~\ref{sec:realeks} was an injected schema-conformant warning, not an AWS-issued FIS reclaim (blocked on an account subscription, not the design), though NTH cannot distinguish the two, so the drain-to-resume path is genuinely exercised; 
(b) predictive's zero wasted compute in Section~\ref{sec:predpay} assumes a configured lead, and Section~\ref{sec:leadtime} shows a small lead suffices, so this is a tuning parameter, not an oracle; 
(c) periodic, an ML-free baseline, matches predictive on wasted compute and wins on makespan and checkpoint cost at the fastest rate (Table~\ref{tab:bench}), and predictive's clean advantage there is zero wasted compute; recovery time is not a reliable comparison at this rate, since only one of five runs completes; and 
(d) the risk model is advisory and trained on a proxy label, so its ceiling is label quality (Section~\ref{sec:riskmodel}). Scope: the benchmark uses a synthetic job with locally injected interruptions, and there is no large-model-scale validation; both are future work, stated as scope, not a claim.

\section{Related Work and Conclusion}

Spot-resilient training systems use elastic rescaling, pipeline templates, or redundant computation~\citep{varuna, bamboo, oobleck} to make training itself survive preemption. Argus does not propose any new resilient-training mechanism but instead asks the orthogonal policy question of when the free reactive notice is beaten by a predictive checkpoint. Checkpoint-restart systems on top of the OS-level primitive~\citep{criu} cut the cost or latency of the checkpoint itself~\citep{checkfreq, gemini}. Node or job failure is learned by ML-for-systems failure and interruption prediction~\citep{node-failure-pred}. For Argus, the risk model is trained on a proxy label (Section~\ref{sec:riskmodel}) and is advisory, thus instead of reporting a strong-predictor claim, we report a benchmark-and-guideline result. Argus builds on Kubernetes operators, which are the standard way to automate stateful workloads~\citep{k8s-operators}.

Argus survives a real Spot drain on real EKS, and an 80-trial benchmark maps when prediction pays: periodic is a strong ML-free baseline, and predictive's advantage is zero wasted compute, clean only when its lead matches the interruption rate. A lead-time sweep turns this into a guideline: a small lead suffices for zero waste, while excess lead over-migrates. Future work includes FIS-forced reclaims, forward interruption-label capture to replace the proxy label, and multi-node large-model validation.

\bibliographystyle{plainnat}
\bibliography{references}

\appendix
\section{Technical appendices and supplementary material}
This appendix holds the supporting artifacts and full benchmark data referenced in the body.

\subsection{Real-EKS interruption survival}

\begin{table}[h]
  \centering
  \caption{Real-EKS Spot interruption timeline.}
  \label{tab:obj1}
  \small
  \begin{tabular}{ll}
    \toprule
    Time (UTC) & Event \\
    \midrule
    12:26:18 & NTH receives \texttt{EC2 Spot Instance Interruption Warning} \\
    12:26:19 & Drain evicts pod \texttt{cifar10-test} (graceful SIGTERM) \\
    12:26:19 & SIGTERM handler writes checkpoint (epoch 8) to S3 \\
    12:26:25 & Node cordoned and drained in 10\,s \\
    $\sim$75\,s later & Replacement pod on healthy node: ``Resuming from epoch 8'' \\
    \bottomrule
  \end{tabular}
\end{table}

\begin{figure}[h]
  \centering
  \includegraphics[width=\linewidth]{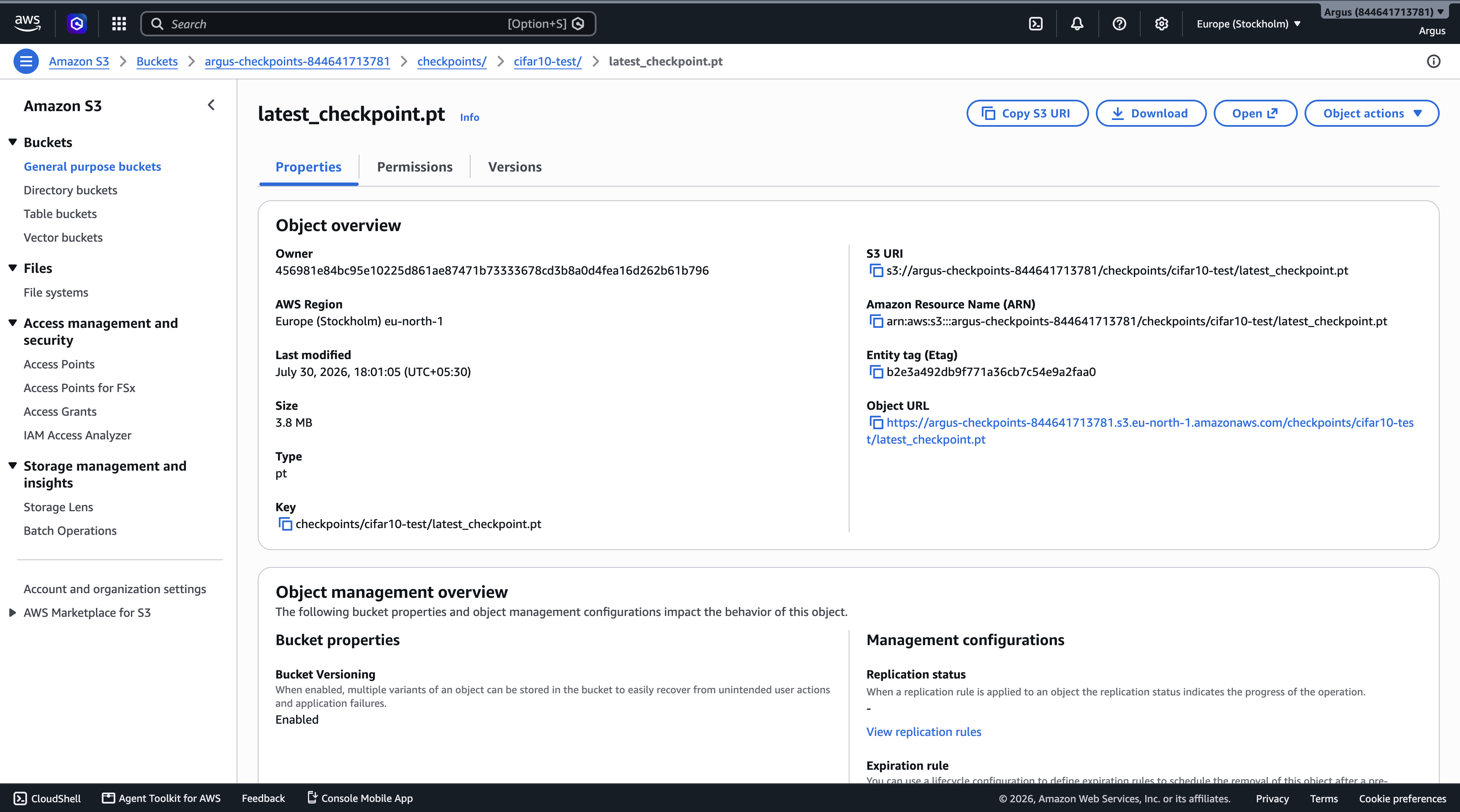}
  \caption{The 3.8\,MB \texttt{latest\_checkpoint.pt} written to the real S3 bucket during the Spot drain (Table~\ref{tab:obj1}), direct evidence the checkpoint survived the interruption, not a simulation.}
  \label{fig:s3ckpt}
\end{figure}

Table~\ref{tab:obj1} is the full drain-to-resume timeline; Figure~\ref{fig:s3ckpt} shows the checkpoint object persisted to the real S3 bucket during the drain.

\subsection{Full benchmark results}

\begin{table}[h]
  \centering
  \caption{Lead-time sensitivity (predictive arm, 20\,s mean interruption
  interval, 5 repetitions). Periodic baseline: 4.3\,s wasted, 78.1\,s
  makespan, 29.0 checkpoints.}
  \label{tab:leadsweep}
  \small
  \begin{tabular}{rrrr}
    \toprule
    Lead (s) & Wasted (s) & Makespan (s) & Checkpoints \\
    \midrule
    2  & 0.0 & 73.1  & 5.4  \\
    5  & 0.0 & 76.2  & 7.4  \\
    10 & 0.0 & 78.0  & 8.8  \\
    15 & 0.0 & 83.3  & 12.6 \\
    20 & 0.0 & 87.0  & 15.2 \\
    30 & 0.0 & 96.7  & 21.8 \\
    45 & 0.0 & 131.7 & 46.6 \\
    60 & 0.0 & 218.3 & 93.4 \\
    \bottomrule
  \end{tabular}
\end{table}

\begin{figure}[h]
  \centering
  \includegraphics[width=\linewidth]{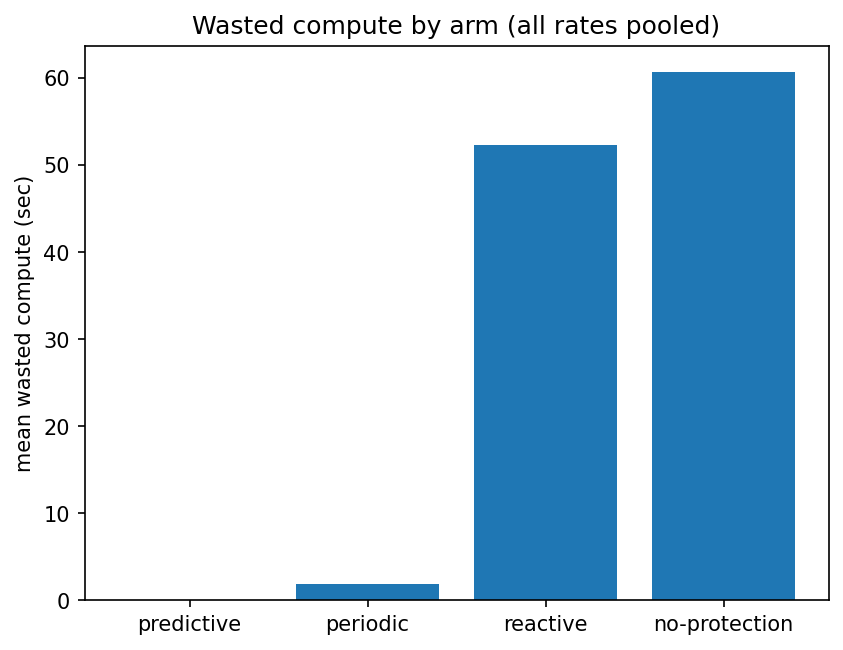}
    \caption{Wasted compute by arm across all four tested interruption rates.}
  \label{fig:waste}
\end{figure}

Figure~\ref{fig:waste} shows wasted compute by arm pooled across all four rates; Table~\ref{tab:leadsweep} gives the complete per-lead data behind Figure~\ref{fig:leadsweep}.

\subsection{Observability}

\begin{figure}[h]
  \centering
  \includegraphics[width=\linewidth]{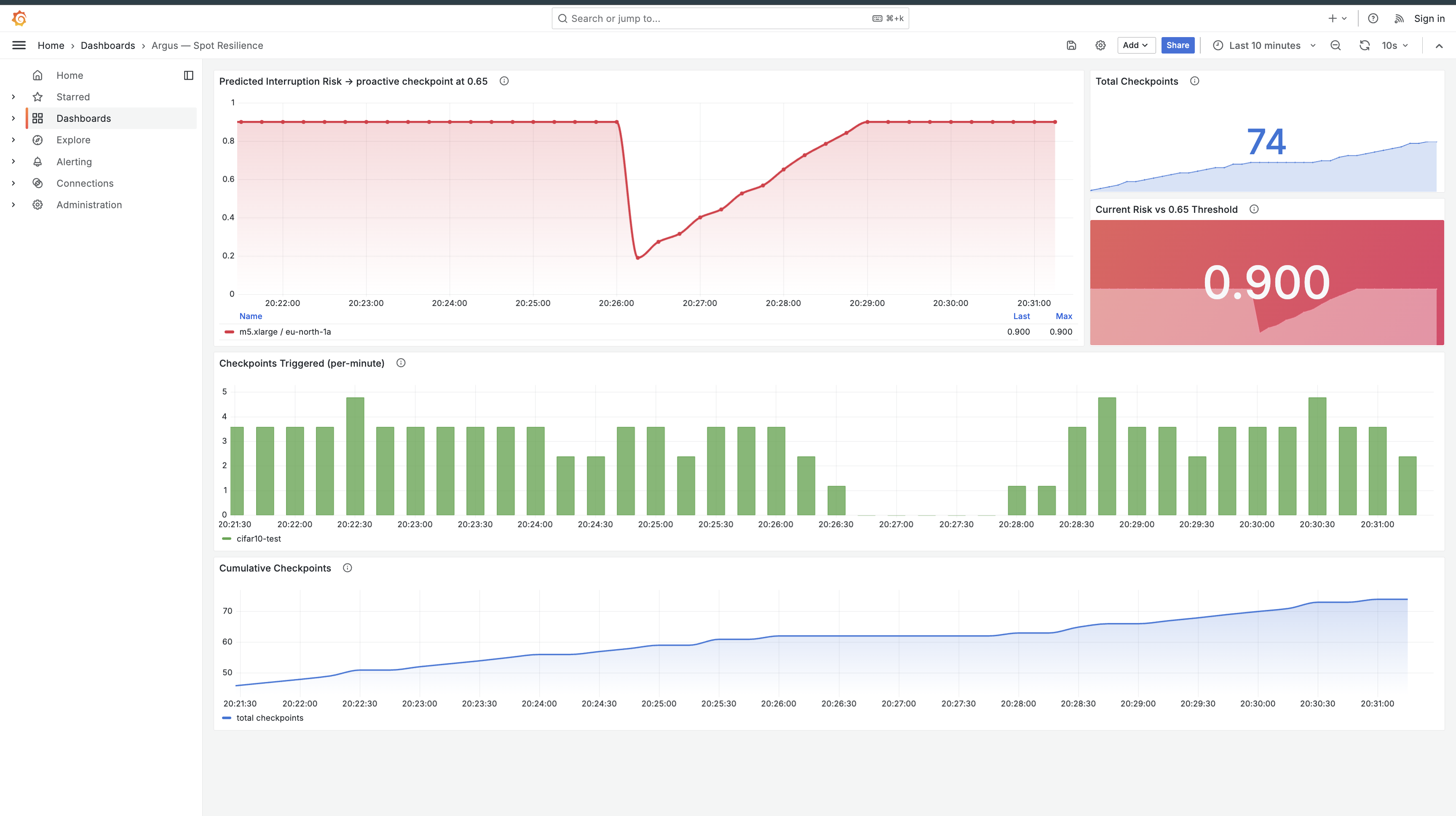}
  \caption{Argus observability: predicted risk crosses the trigger threshold and
  a proactive checkpoint fires at that instant (real operator, real S3/SQS).}
  \label{fig:grafana}
\end{figure}

Figure~\ref{fig:grafana} is the live Grafana dashboard.

\end{document}